\documentclass[10pt,twocolumn,letterpaper]{article}
\usepackage{wacv}
\usepackage{times}

\usepackage{graphicx}
\usepackage{comment}
\usepackage{amsmath,amssymb} % define this before the line numbering.
\usepackage{color}

\usepackage{multirow}
\usepackage{booktabs}

\usepackage{epsfig}
\usepackage{adjustbox}
\usepackage{subfigure}

\wacvfinalcopy % *** Uncomment this line for the final submission

\ifwacvfinal
\def\assignedStartPage{1} % *** Enter the assigned starting page number (instead of 9876)
\fi

\ifwacvfinal
\usepackage[breaklinks=true,bookmarks=false]{hyperref}
\else
\usepackage[pagebackref=true,breaklinks=true,colorlinks,bookmarks=false]{hyperref}
\fi

\ifwacvfinal
\else
\fi

\begin{document}

\title{Deep Reinforcement Learning with Label Embedding Reward\\for Supervised Image Hashing} % Replace with your title

\author{
Zhenzhen Wang\\
Nanyang Technological University, Singapore\\
{\tt\small zwang033@e.ntu.edu.sg}
\and
Weixiang Hong\\
Ant Financial Services Group, Alibaba Inc.\\
{\tt\small hwx229374@antfin.com}
\and
Junsong Yuan\\
State University of New York at Buffalo, USA\\
{\tt\small jsyuan@buffalo.edu}
}
%******************
\maketitle

\begin{abstract} 
Deep hashing has shown promising results in image retrieval and recognition. Despite its success, most existing deep hashing approaches are rather similar: either multi-layer perceptron or CNN is applied to extract image feature, followed by different binarization activation functions such as sigmoid, tanh or autoencoder to generate binary code. 
In this work, we introduce a novel decision-making approach for deep supervised hashing. We formulate the hashing problem as travelling across the vertices in the binary code space, and learn a deep Q-network with a novel label embedding reward defined by Bose-Chaudhuri-Hocquenghem (BCH) codes to explore the best path. Extensive experiments and analysis on the CIFAR-10 and NUS-WIDE dataset show that our approach outperforms state-of-the-art supervised hashing methods under various code lengths.
\end{abstract}
	
\section{Introduction}
	
	Binary embedding, \textit{a.k.a.} hashing, has attracted much attention in recent years due to the rapid growth of image and video data on the web \cite{weiss2009spectral,gong2011iterative,shen2015supervised,Liong_2015_CVPR,yang2017supervised,Duan_2017_CVPR,cao2017hashnet,NIPS2018_7360}. Generally speaking, binary embedding aims to encode high-dimensional image features into compact binary codes while preserving their pairwise similarities. Due to the storage efficiency and low computational cost of compact binary codes, hashing has become one of the most popular techniques for image and video search.
	
	\begin{figure}[t]
		\begin{center}
			\includegraphics[width=0.7\linewidth]{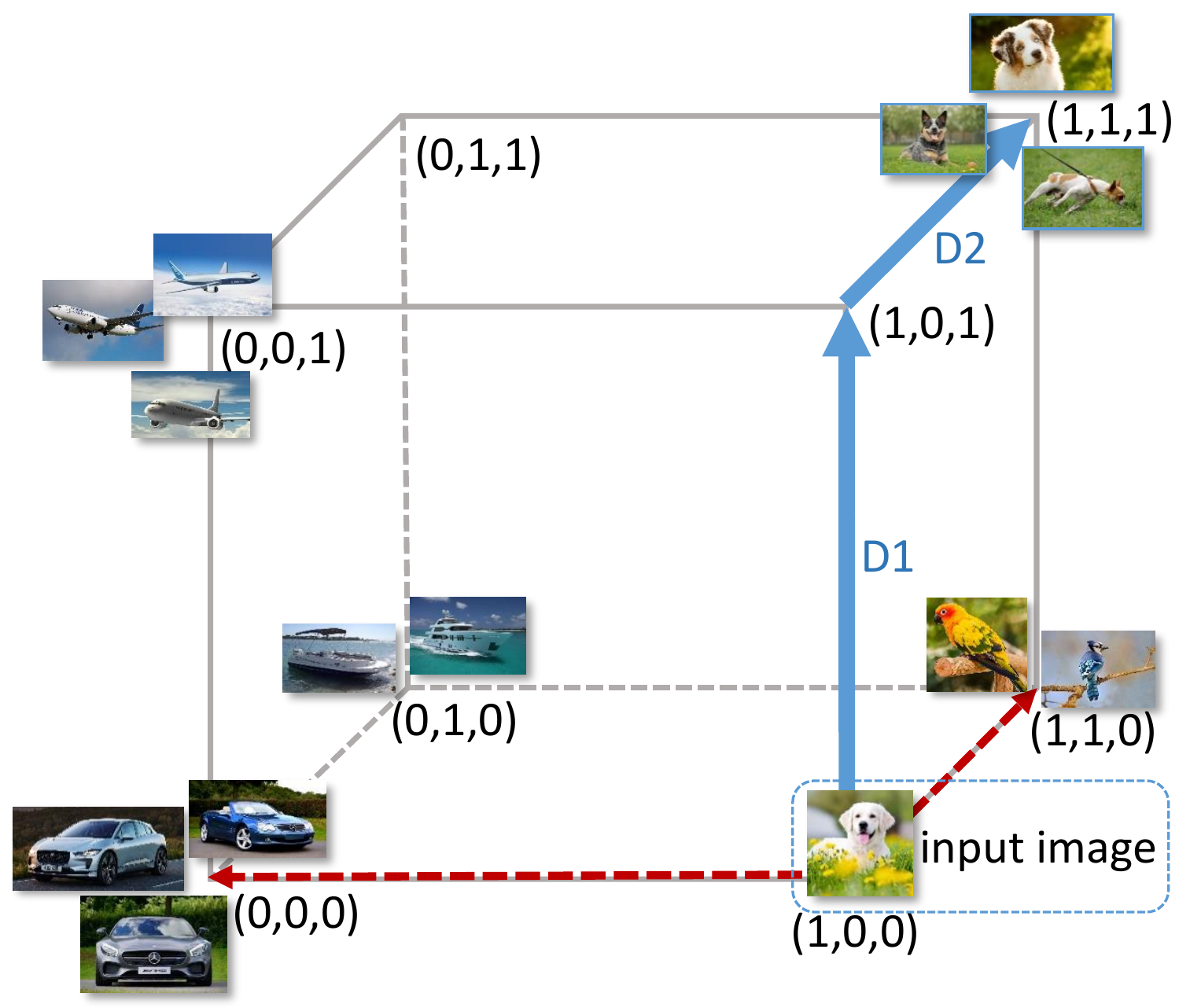}
		\end{center}
	%\vspace{-2mm}
		\caption{Illustration of learning to hashing by travelling in the binary space. For each image feature, we utilize a decision-making network to determine which direction to go in each step, thus the binary embedding is converted to a sequential decision making process. Images of different classes are located to different vertices in the binary spaces. We use blue arrows to show the Decision 1 and Decision 2 of the best path to embed the input image, red arrows to denote undesired directions to proceed.}
		\label{figure_illustration}
	\end{figure}
	
	Existing hashing methods can be roughly grouped into two categories: traditional methods and deep hashing. Traditional methods \cite{weiss2009spectral,gong2011iterative,shen2015supervised} take a feature vector as input and embed it into a compact binary vector, while deep hashing \cite{Liong_2015_CVPR,yang2017supervised,Duan_2017_CVPR,cao2017hashnet} jointly optimizes the feature extraction and binary embedding. Although deep hashing, especially deep supervised hashing, has demonstrated superior retrieval performance over traditional hashing methods, most existing deep hashing methods suffer from the quantization loss as well as the train/test inconsistency. For example, during training, either multi-layer perceptron \cite{Liong_2015_CVPR} or CNN \cite{yang2017supervised,Duan_2017_CVPR,cao2017hashnet} is used to extract image feature, which is then approximately binarized by activation units such as sigmoid \cite{zhao2015deep,yang2017supervised}, tanh \cite{cao2017hashnet}, or autoencoder \cite{Duan_2017_CVPR}. During inference, these methods typically round the outputs of the activation unit by a rigid sign function to generate the binary code. 
	
	In this work, we introduce a novel decision-making approach for deep supervised hashing. We propose to cast the supervised image hashing problem as a Markov Decision Process (MDP) \cite{sutton1998reinforcement}, and solve the MDP by deep reinforcement learning. As shown in Figure~\ref{figure_illustration}, we regard the binary code of an input image as a vertex in the binary code space, and consider the binary hashing problem as the process of traveling across the vertices in the binary space, from an initial location to the desired location. Specifically, given an initial binary code for an input image, the agent learns to execute a sequence of actions to move the initial code to a better location in the binary code space. The policy for the agent to decide which action to take is learned by deep reinforcement learning to maximize the accumulative rewards. In testing phase, the agent moves the input image from current position to next position following the learned policy without receiving rewards anymore. To stimulate the proposed agent, we first encode the image label into the large-margin error-correcting codes, \textit{e.g.}, Bose-Chaudhuri-Hocquenghem (BCH) codes \cite{hocquenghem1959codes,bose1960class}, then design a reward function to measure how well the current binary embedding does in the search process. We incorporate the reward function in a reinforcement learning setting to learn a binary embedding policy, based on the Deep Q-Network \cite{mnih2015human}. As a result, the trained agent can efficiently localize an input image to desired position in the binary code space.
	
	We term our approach as DRLH for Deep Reinforcement Learning based Hashing, which is fundamentally different from prior arts. First, our proposed DRLH does not require quantization from real-value representation into binary code, thus it does not suffer from the quantization loss as well as the inconsistency between training and testing phases. Moreover, compared with traditional methods \cite{weiss2009spectral,gong2011iterative}, our method can benefit from the powerful learning ability of neural network and is an end-to-end optmization process. Last, in contrast to existing deep hashing approaches, our method does not rely on the rigid sign function to generate binary code, thus does not require any relaxization to enable gradient descent.
	
	We conduct detailed analysis on DRLH to evaluate its merits and properties. Extensive experiments on the CIFAR-10 dataset \cite{krizhevsky2009learning} and NUS-WIDE dataset \cite{chua2009nus} show that the proposed method performs favourably against state-of-the-art unsupervised and supervised hashing approaches, including HashNet~\cite{cao2017hashnet}, DSDH~\cite{li2017deep}, SSDH~\cite{yang2017supervised}, and GreedyHash~\cite{NIPS2018_7360}, \textit{etc}. The contributions of this paper are summarized as following: 
	
	\begin{itemize}
		\item We present a novel decision-making approach DRLH for deep supervised hashing utilizing a deep Q-network. It directly operates binary bits by conducting decision-making process, without relaxing the binary constraint.
		
		\item To learn our Q-network, we introduce BCH codes to embed the image labels into the binary spaces with large margin. Our experiments suggest that as an instance of the error correcting code, BCH codes work well for constructing the reward function.
		
		%\item Our method achieves state-of-the-art performance on the CIFAR-10 dataset \cite{krizhevsky2009learning} and NUS-WIDE dataset \cite{chua2009nus}. 
		
	\end{itemize}
	
	\section{Related Work}
	
	\subsection{Binary Embedding}
	\label{Subsection_Related_work_Binary_embedding}
	
	Binary embedding is a promising technique for large-scale information retrieval. Compact binary code enables efficient linear scan for two reasons: (1) computing Hamming distances (with hardware support) is much faster than computing distances between high-dimensional floating-point vectors; and (2) the entire database consumes much smaller memory, so it may reside in fast memory rather than hard disk. Existing hashing methods could be grouped into two categories: traditional methods and deep hashing.
	
	Traditional methods \cite{weiss2009spectral,gong2011iterative,shen2015supervised} take a feature vector as input and output a compact binary vector. Since it is NP-hard to directly learn the optimal binary code \cite{weiss2009spectral}, most traditional methods first relax the discrete constraint in training phase, then round the continuous solutions to obtain the binary code in testing phase. However, such an inconsistency between training and testing could result in undesired performance drop. Though some recent hashing methods \cite{shen2015supervised} can be directly trained under discrete constraints, they still need to perform binarization in testing phase, thus still suffer from the quantization loss as well as the train/test inconsistency. 
	
	Moreover, since the binary embedding procedure is independent to the feature extraction stage, the performance of traditional hashing methods is constrained by the quality of original features. To address this problem, deep hashing \cite{Liong_2015_CVPR,yang2017supervised,Duan_2017_CVPR,cao2017hashnet} is recently proposed to jointly optimize the feature extraction and binary embedding steps. Typically, most deep hashing methods first adopt multi-layer perceptron \cite{Liong_2015_CVPR} or convolutional neural network \cite{yang2017supervised,Duan_2017_CVPR,cao2017hashnet} to extract the real-value representation of input image, then quantize the real-value representation to binary code. Thanks to the simultaneous optimization of feature extraction and binary embedding steps, deep hashing, especially deep supervised hashing, has demonstrated superior retrieval performance than those traditional hashing methods.
	
	Nevertheless, the train/test inconsistency in deep hashing is even more severe than in traditional hashing, due to the inherent conflict between hashing and back propagation. For example, a rigid sign function, whose gradient is either zero or does not exist, is often necessary for binarizing the real-value representation to binary code. Meanwhile, deep neural networks require all components to be differentiable, so that the parameters of network could be gradually updated by back-propagation. To enable the backward pass of network in the training phase, existing deep hashing methods either approximate the rigid sign function by smooth functions such as sigmoid \cite{zhao2015deep,yang2017supervised} or tanh \cite{cao2017hashnet}, or directly encourage the activation close to zero or one by applying a penalty \cite{li2016feature,wang2016deep,li2017deep}. After the training is completed, the rigid sign function is harnessed to output the binary code. Such an inconsistency could lead to sub-optimal performance. Also, sigmoid or tanh unit is known inappropriate for deep learning, as the gradients of them are nearly zero for most inputs. Tiny gradient could cause gradient vanishing, which is exactly the motivation of ReLU \cite{jarrett2009best}. 
	
	\subsection{Decision Process}
	
	Decision process \cite{sutton1998reinforcement} is the core problem in computer gaming \cite{silver2017mastering}, control theory \cite{mnih2015human}, navigation \cite{zhu2017target} and path planning \cite{Zhu_2017_ICCV}, \textit{etc}. In those problems, there exist agents that interact with the environment, execute a series of actions, and aim to fulfill some pre-defined goals. Reinforcement learning \cite{konda2000actor}, known as ``a machine learning technique concerning how software agent ought to take actions in an environment so as to maximize some notion of cumulative reward'', is well suited for the task of decision process. Recently, professional-level computer Go program was designed using deep neural networks and Monte Carlo Tree Search \cite{silver2017mastering}. Human-level gaming control was achieved through deep Q-learning \cite{mnih2015human}. A visual navigation system was proposed recently to make sequential decision based on successor representation \cite{Zhu_2017_ICCV}. In this paper, we propose to formulate binary embedding as a Markov decision process and solve it with deep reinforcement learning \cite{konda2000actor}.
	
	\subsection{Binary Embedding by Decision Process}
	\label{Subsection_Related_work_Binary_Embedding_by_Decision_Process}
	
	Binary embedding is often addressed under unsupervised or supervised learning settings \cite{wang2018survey}, and is rarely formulated as decision process and solved by reinforcement learning. In recent, some studies attempts to introduce reinforcement learning techniques for obtaining binary code.
	
	Specifically, \cite{Duan_2018_CVPR} uses decimal numbers as bit representation, then quantizes them to binary code by rigid sign function, where DRL is exploited to increase the reliability of quantization. Yuan \textit{et al.} \cite{Yuan_2018_ECCV} propose a deep hashing CNN to generate binary code, and resort to policy gradient method from reinforcement learning theory to optimize their deep hashing network. Zhang \textit{et al.}\cite{zhang2018deep} make one step further by treating a batch of images as environment, triplet-loss as reward, and adjusting a combination of hashing function to maximise the reward in reinforcement learning. Unluckily, the hard binary constraint in \cite{zhang2018deep} is compromised, and train/test inconsistency still exists. Different from these methods, our approach directly operates binary code, there is neither decimal representation nor quantization loss. As shown in Section~\ref{Compare_with_deep_hashing}, our proposed approach surpasses these methods under various settings.
	
	\section{Proposed Approach}
	\label{Section_Proposed_Approach}
	
	Since Markov decision process (MDP) offers a formal formulation of madeling an agent that makes sequential decisions, we propose to cast our problem of binary embedding into the framework of MDP \cite{sutton1998reinforcement}. Specifically, our formulation considers the binary space as the environment, where the agent moves an image towards the desired vertex using a set of actions. Similar to existing supervised hashing works \cite{shen2015supervised,cao2017hashnet,zhang2016efficient,wang2016deep}, the goal of our agent is to generate a good embedding for each image that minimizes the Hamming distance among those images of the same class, and maximizes the Hamming distance among those images of different classes. The agent also has a state representation recording the current binary code and past actions, and receives corresponding rewards for each decision made during the training phase. In testing, the agent neither receives rewards, nor updates the model. It just follows the learned policy to produce the binary codes.
	
	Formally, a MDP contains a set of states $\mathcal{S}$, a set of actions $\mathcal{A}$ and a reward function $\mathcal{R}$. We elaborate details of each component in this section, and present technical details of the learning process in Section~\ref{sect4: DRL}.
	
	\subsection{Action}
	\label{Subsection_Action}
	
	Given an initial binary code of $b$ bits, the set of actions $\mathcal{A}$ consists of $b$ transformations that can be applied to the current binary embedding and one action to terminate the search process, thus $|\mathcal{A}| = b + 1$. The design of the action set is quite straightforward for conducting binary embedding, \textit{i.e.}, the $k$-th action in $\mathcal{A}$ is to flip the $k$-th bit of the current binary code. In other words, the $k$-th action always flips the $k$-th bit between $0$ and $1$. In addition, we define an action that terminates the sequence of current search, which will be triggered once the current binary code of an image has been located in the desired destination.
	
	\subsection{State}
	\label{Subsection_State}
	
	The state representation at step $t$ is a tuple $s_t = (f, e_t, h_t)$, where $f$ is the feature vector of the input image, $e_t$ is the current binary code of the input image, and $h_t$ is a vector that records the history of actions taken. The feature vector $f$ for the input image could be either traditional representation such as GIST~\cite{oliva2001modeling}, or the responses of a CNN model as in recent hashing methods \cite{Duan_2017_CVPR,cao2017hashnet}. We will experimentally show the differences in Section~\ref{Section_Experiments}. The binary code $e_t$ indicates the current location in the binary code space. The ultimate goal of the agent is to move the input image to a desired location in the binary code space by executing a sequence of actions. We initialize $e_0$ for each image as a random binary code. The history $h_t$ is a binary vector that records which actions have been taken in the past. Given the set $\mathcal{A}$ of all possible actions, each action in $h_t$ is represented by a $b$-dimensional one-hot binary vector, whose all values are zero except for the one corresponding to the taken action. The terminate action will not be recorded. The history vector records $10$ past actions, \textit{i.e.}, $h_t \in \{ 0, 1 \} ^ {10 \times b}$. The history $h_t$ carries the information about what has happened before time $t$, which is helpful to stabilize the search trajectories that might get stuck in repetitive cycles \cite{mnih2015human}.
	
	\subsection{Reward Function}
	\label{Subsection_Reward_Function}
	
	In deep reinforcement learning, reward function is the key to guide the agent in the exploration process. In most gaming problems such as Go \cite{silver2017mastering}, defining the reward is easy, \textit{i.e.} win or lose. In navigation \cite{Zhu_2017_ICCV}, the reward is reaching the target or not. However, the reward in hashing is not so straightforward, since we only know the labels of the training images, instead of the ``ground-truth binary codes''.
	
	Recall our discussion in the beginning of Section~\ref{Section_Proposed_Approach}, the goal of our agent is to generate a binary embedding for each image that minimizes the Hamming distance to those images of the same class, and maximizes the Hamming distance to those images of different classes. Such a design principle is partially inspired by spherical hashing \cite{heo2012spherical} and center loss \cite{wen2016discriminative}. Specifically, spherical hashing \cite{heo2012spherical} proposes to group similar images into the same hyperball in the binary space, while center loss \cite{wen2016discriminative} aims to learn the image representations that are compact for the same class. 
	
	Following the above principle, we propose to encode $C$ different classes in the training set into the $C$ binary codewords of $b$ bits, such that the distance between any two codewords is as large as possible. Such kind of codes is known as linear code \cite{mackay2003information}. Specifically, a code has minimum distance $\mathrm{D}$ if $\mathrm{D}$ is the largest integer such that the distance between any distinct codewords is at least $\mathrm{D}$. Then, the balls of radius $\mathrm{R} = (\mathrm{D}-1)/2$ centered around the codewords do not overlap, and the code is said to be a $[(\mathrm{D}-1)/2]$-error-correcting code. In ideal case where these balls tile the whole space, the code is called as perfect code. Unfortunately, there is no trivial perfect codes for binary codes \cite{krotov2008number}, hence we propose to build our reward function based on the BCH codes \cite{bose1960class}, which could encode data samples into large-margin codewords. As the length of BCH codes must be $2^m - 1$ such that $m \geq 3$, we could obtain the binary code of desired length by dropping the redundant bits or padding some random bits. For example, if the desired code length $b$ is $16$, we can first generate BCH codes with $m = 4$ , then pad $1$ random bit to the BCH codes to produce the desired binary code.
	
	At state $\mathcal{S}_t$, we use $d_{pos}^{t}$ to denote the distance between the current binary code and the code of ground-truth class, $d_{neg}^{t}$ to denote the average distance between the current binary code and the code of other classes. In case of multi-label dataset like NUS-WIDE \cite{chua2009nus}, $d_{pos}^{t}$ is computed using the binary embedding of the nearest ground-truth class at time step $t$. Then, the reward function is estimated using the margin of the distance from one state $t$ to another $t'$, \textit{i.e.}, 
	\begin{equation}
	\label{Equation_reward_definition}
	\mathcal{R}_a(\mathcal{S}_t, \mathcal{S}_{t'}) = \big( d_{pos}^{t} - d_{neg}^{t} \big) - \big( d_{pos}^{t'} - d_{neg}^{t'} \big).
	\end{equation}
	
	Intuitively, Equation~(\ref{Equation_reward_definition}) encourages the agent to move towards the ground-truth class and away from other classes. The reward $\mathcal{R}_a(\mathcal{S}_t, \mathcal{S}_{t'})$ is granted to the agent when it chooses the action $a$ to move from state $\mathcal{S}_t$ to $\mathcal{S}_{t'}$. The agent is penalized for taking the image away from the target vertex, and is rewarded for moving the input image towards the desired location. Once no action further decreases the distance between the current binary embedding and target embedding, the termination action is triggered to stop further movement. Note that the reward function is only used during the training phase, in testing phase, the agent just follows the learned policy.
	
	Since the termination trigger does not change current binary code, the differential of distance in Equation~\ref{Equation_reward_definition} will always be zero. Thus, we further define a different reward function for the termination trigger as following:
	\begin{equation}  
	\label{Equation_reward_trigger}
	\mathcal{R}_{w}(\mathcal{S}_t, \mathcal{S}_{t'}) = 
	\left\{  
	\begin{array}{lr}  
	+ \sigma, \;\text{if}\; d_{pos}^{t} \leq \eta, \\  
	- \sigma, \;\text{otherwise}, &    
	\end{array}  
	\right.  
	\end{equation}
	where $w$ is the termination action, $\sigma$ is the termination reward, set to $5.0$ in our experiments, and $\eta$ is a threshold indicating the minimum distance allowed to treat the final binary code as a positive output.
	
	Finally, we note that the number of steps is regarded as a cost by the agent due to the discount factor involved in Q-learning. Thus, the agent is supposed to take the short path to move the input image since any unnecessary step will decrease the reward received.
	
	\section{Deep Reinforcement Learning}
	\label{sect4: DRL}
	
	In each episode, the goal of the agent is to move the binary embedding towards target location by performing a sequence of action, so that the sum of the received rewards are maximized. Therefore, we aim to find a policy function $\pi(s_t)$ that consumes a state representation $s_t$ as input, and guides the agent on selecting the next action $a_{t+1}$. As the state space is very large and the reward function is data-dependent, we utilize Q-learning \cite{sutton1998reinforcement} to tackle this problem. Similar to \cite{mnih2015human}, we estimate the action-value function with a neural network and replay-memory mechanism. Upon the action-value function $Q(s, a)$ is learnt, the agent can traverse the binary space by simply selecting the action $a$ with the maximum estimated value, \textit{i.e}., $\pi(s_t)=\arg\max_{a} Q(s_t, a)$.
	
	\subsection{Q-network Implementation}
	
\begin{figure}[!t]
		\begin{center}
			\includegraphics[width=0.8\linewidth]{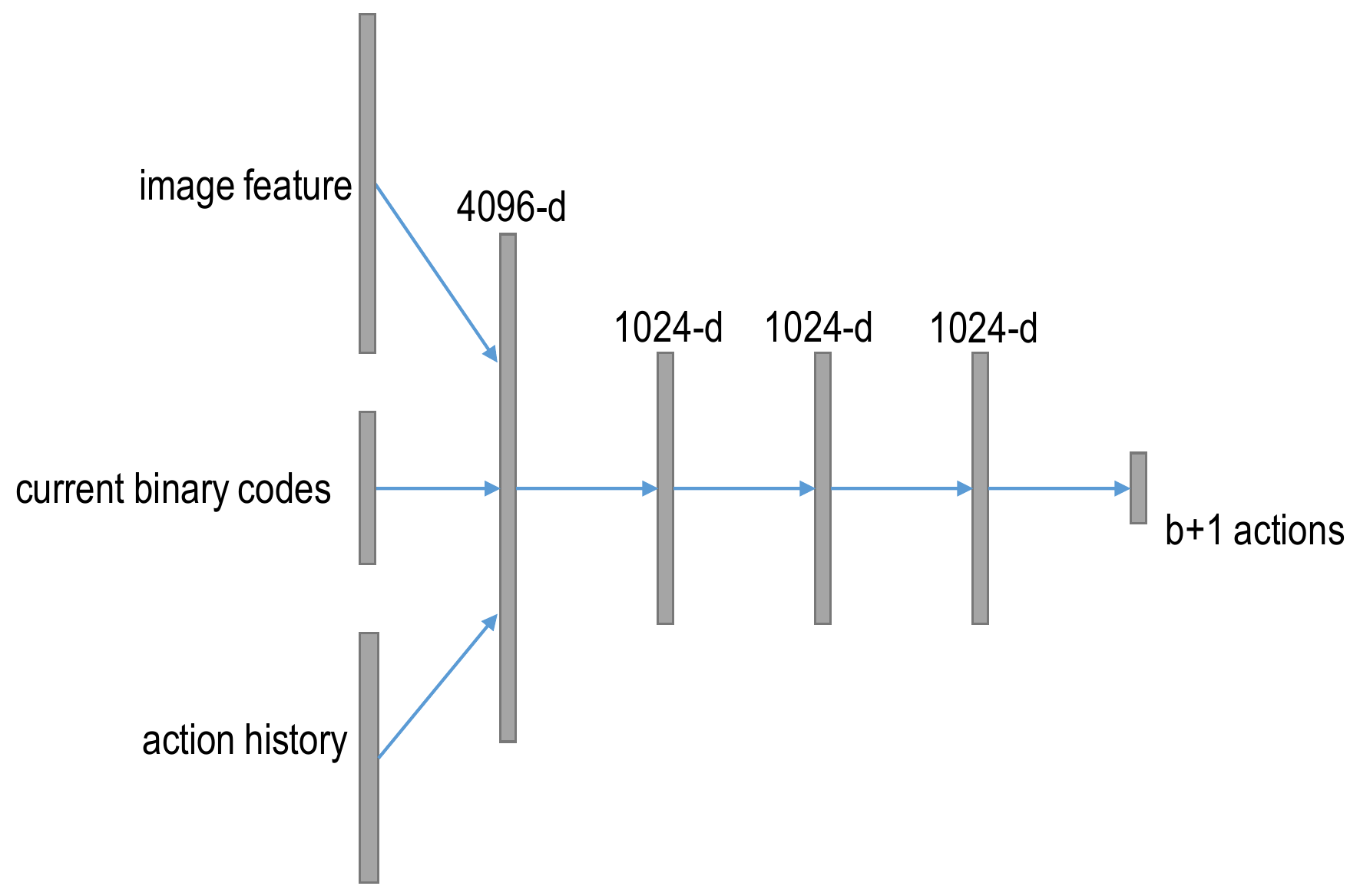}
		\end{center}
		%\vspace{-2mm}
		\caption{Architecture of the proposed Q-Network. The input image feature could be represented by traditional feature such as GIST~\cite{oliva2001modeling} or deep feature extracted from CNN model. The image representation is concatenated with the current binary code and action history vector to compose the state representation, which is processed by the Q-network to predict the probabilities for the $b+1$ actions.}
		\label{figure_network}
		%\vspace{-2mm}
	\end{figure}

	We use a Q-network which consumes the state representation discussed in Section~\ref{Subsection_State} and outputs the value of the $|\mathcal{A}| = b + 1$ actions presented in Section~\ref{Subsection_Action}. The architecture of our Q-networks is illustrated in Figure~\ref{figure_network}. We apply dropout to the responses of each fully-connected layer.
	
	\subsection{Learning Details}
	
	Our deep Q-network is randomly initialized, then updated with stochastic gradient descent (SGD). In the training phase, we harness the $\epsilon$-greedy strategy \cite{sutton1998reinforcement} to gradually shift from exploration to exploitation by adjusting the value of $\epsilon$. Since the exploration in the binary code space does not proceed with random actions as in most existing reinforcement learning methods, we follow \cite{caicedo2015active} to adopt a guided exploration strategy with apprenticeship learning \cite{coates2008learning}, which is based on the demonstrations made by an expert to the agent.
	
	% We initialize our deep Q-network randomly, and update them using stochastic gradient descent. We utilize $\epsilon$-greedy strategy in the training, which gradually shifts from exploration to exploitation according to the value of $\epsilon$ \cite{sutton1998reinforcement}. Since the exploration in the binary code space does not proceed with random actions like most existing reinforcement learning works, we follow \cite{caicedo2015active} to use a guided exploration strategy with apprenticeship learning \cite{abbeel2004apprenticeship, coates2008learning, levine2016end}, which is based on demonstrations made by an expert to the agent.
	
	We train our Q-network with the $\epsilon$-greedy strategy for $25$ epochs. The agent interacts with all training images in each epoch. During the first $15$ epochs, $\epsilon$ decreases linearly from $1.0$ to $0.1$ to encourage the exploration. After the $15$-th epoch, $\epsilon$ is fixed to $0.1$ so that the agent updates the model parameters according to the experiences produced by its own decisions.
	
	Once an agent is trained with the procedure above, it learns to move the current embedding towards the target vertex. {At each step of testing, the agent selects an action to transform the current code or triggers the terminator once the current code is considered to reach the desired location.} From the perspective of travelling in the binary code space, the perfect agent ought to take at most $b$ steps to reach the best vertex, thus, we set the maximum step $\mathrm{M} = b$ in the testing phase.
	
	\section{Experiments}
	\label{Section_Experiments}
	
	\subsection{Experiment Settings}
	
	We experimentally investigate our DRLH on two public benchmark datasets: CIFAR-10 \cite{krizhevsky2009learning} and NUS-WIDE \cite{chua2009nus}. CIFAR-10 is a dataset containing 60,000 color images in 10 classes, and each class contains 6,000 images with a resolution of $32 \times 32$; NUS-WIDE is a multi-label image dataset containing 269,648 color images in total with 5,018 unique tags. Each image is annotated with one or multiple class labels from the 5,018 tags. Following the common practice, we use a subset of 195,834 images which are associated with the 21 most frequent concepts. Each concept consists of at least 5,000 color images in this dataset.
	
	We roughly divide the existing hashing methods into two groups: traditional hashing methods and deep hashing methods. The compared traditional hashing methods consist of unsupervised and supervised methods. Unsupervised hashing methods include SH~\cite{weiss2009spectral} and ITQ~\cite{gong2011iterative}. Supervised hashing methods include KSH \cite{liu2012supervised}, FastH~\cite{lin2014fast}, LFH~\cite{zhang2014supervised}, and SDH~\cite{shen2015supervised}. The deep hashing methods include GraphBit~\cite{Duan_2018_CVPR}, DSRH~\cite{zhao2015deep}, DQN~\cite{cao2016deep}, NINH~\cite{lai2015simultaneous}, DPSH~\cite{li2016feature}, SSDH~\cite{yang2017supervised}, PGDH~\cite{Yuan_2018_ECCV}, DRLIH~\cite{zhang2018deep}, VDSH~\cite{zhang2016efficient}, DTSH~\cite{wang2016deep}, DSDH~\cite{li2017deep}, SDSH~\cite{Pidhorskyi1_2018_ACCV}, HashNet~\cite{cao2017hashnet}, GreedyHash~\cite{NIPS2018_7360}, TALR-AP~\cite{He2018CVPR}, HBMP~\cite{Cakir2018ECCV} and  DAGH~\cite{Chen2019iccv}. Note that DPSH~\cite{li2016feature}, DTSH~\cite{wang2016deep}, DSDH~\cite{li2017deep}, TALR-AP~\cite{He2018CVPR}, HBMP~\cite{Cakir2018ECCV} and  DAGH~\cite{Chen2019iccv} are based on the CNN-F~\cite{chatfield2014return} architecture, while DQN~\cite{cao2016deep}, DSRH~\cite{zhao2015deep}, SSDH~\cite{yang2017supervised}, HashNet~\cite{cao2017hashnet} and GreedyHash~\cite{NIPS2018_7360} are based on AlexNet~\cite{krizhevsky2012imagenet} architecture. Both CNN-F network architecture and AlexNet architecture consist of five convolutional layers and two fully connected layers. In order to have a fair comparison, most of the results are directly reported from previous works.
	
	Following the common practice, we use mean Average Precision (mAP), \textit{i.e.}, the area under the recall-precision curve of Hamming ranking, to evaluate the image retrieval quality. Similar to \cite{li2017deep}, when computing mAP for NUS-WIDE dataset in Section~\ref{Compare_with_traditional_methods} and \ref{Comparison_with_traditional_methods_using_deep_learned_features}, we only consider the top 5,000 returned neighbors for the convenience of comparison with traditional methods; otherwise, we consider the top 50,000 returned neighbors. All codes and trained models will be made public in future.
	
	\subsection{Ablation Studies}
	
	\subsubsection{Update CNN or not?}
	
\begin{figure}[!t]
		\centering
		\subfigure{\includegraphics[width=4cm]{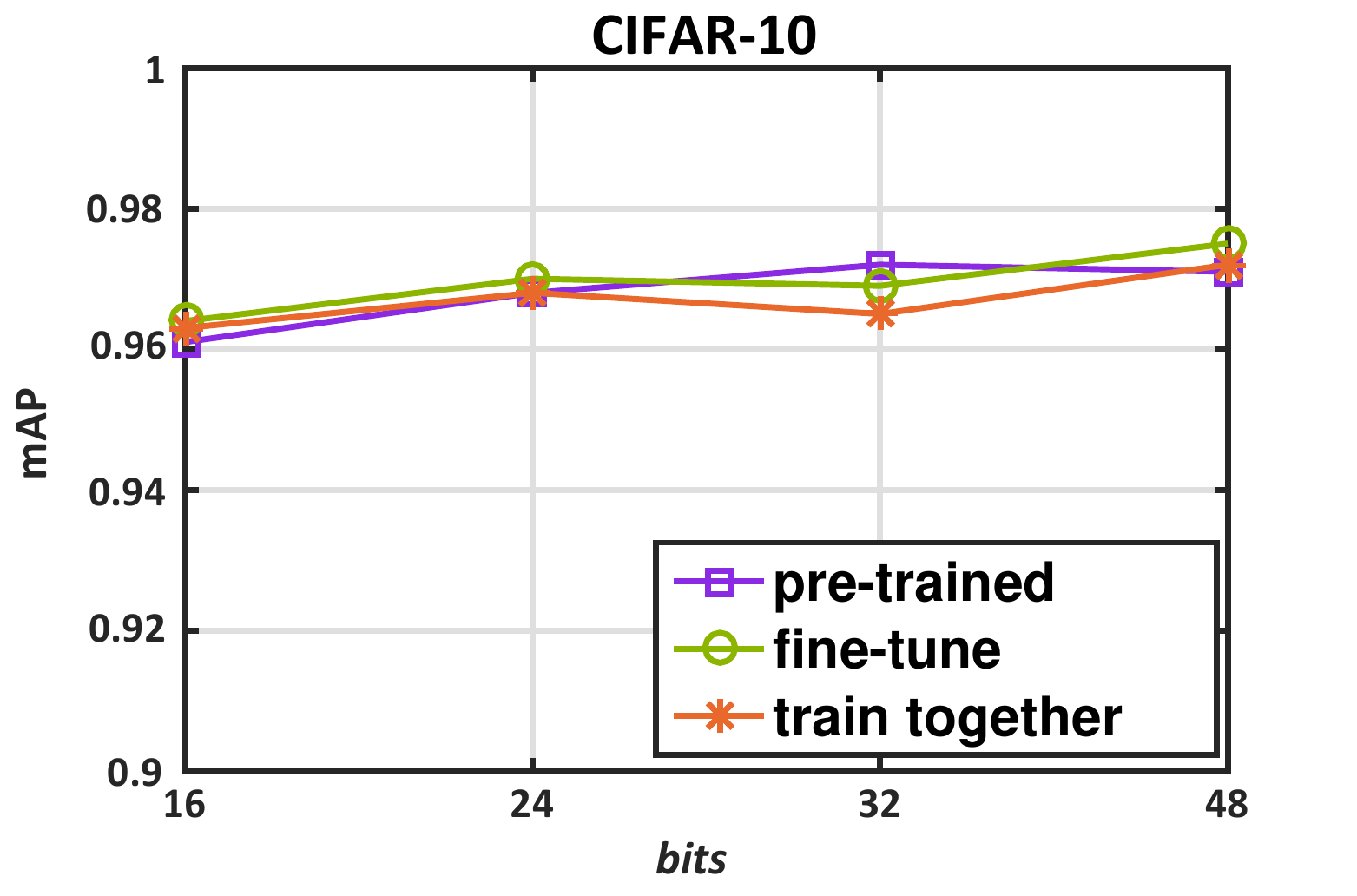}}
		\subfigure{\includegraphics[width=4cm]{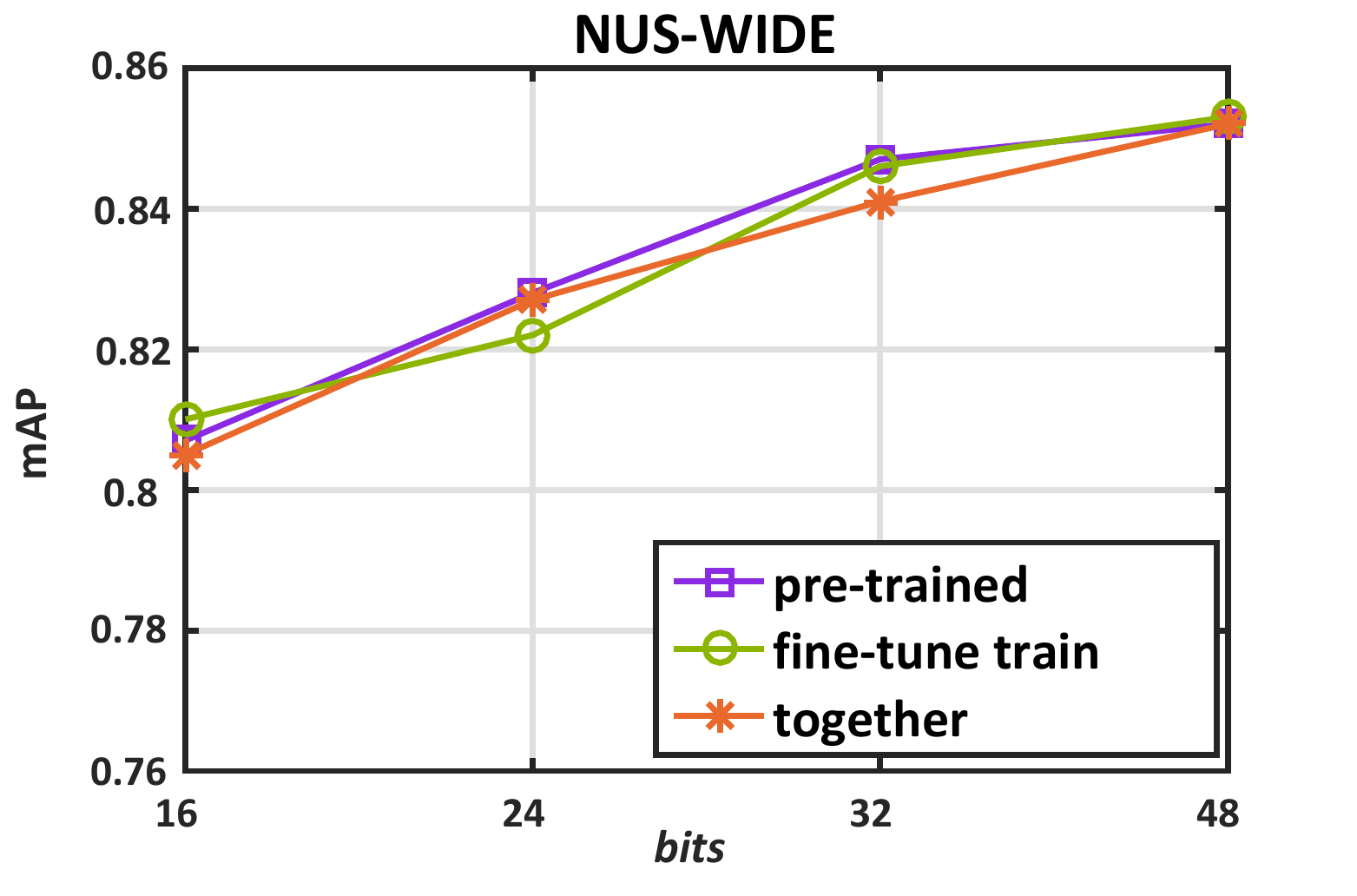}}
		%\vspace{-2mm}
		\caption{mAP for different strategies.}
		\label{fig_updatecnn}
		%\vspace{-2mm}
	\end{figure}

	Besides the pre-trained AlexNet, we also report another two different strategies for computing image representation. One is to finetune the pre-trained AlexNet on the CIFAR-10 dataset, another is to update the CNN weights together with the Q-network. The experimental results are reported in Figure~\ref{fig_updatecnn}. It can be seen that the pre-trained feature performs comparably against the two kinds of fine-tuned features, the possible reasons are: 1) the pre-trained AlexNet is trained on a huge dataset ImageNet, which enables it to capture the generic discriminative features for images in CIFAR-10 dataset; 2) our Q-network is able to diminish the small difference between input features.
	
	Due to the minor differences between the 3 strategies, we simply use the pre-trained AlexNet to extract image feature in the following experiments. In this way, we only update the parameters of Q-network and keep CNN parameters fixed during the training of our RL agent, which is faster than the joint training.
	
	%\subsubsection{Action Set Design}
	
\begin{figure}[!t]  
		\centering
		\subfigure{\includegraphics[width=4cm]{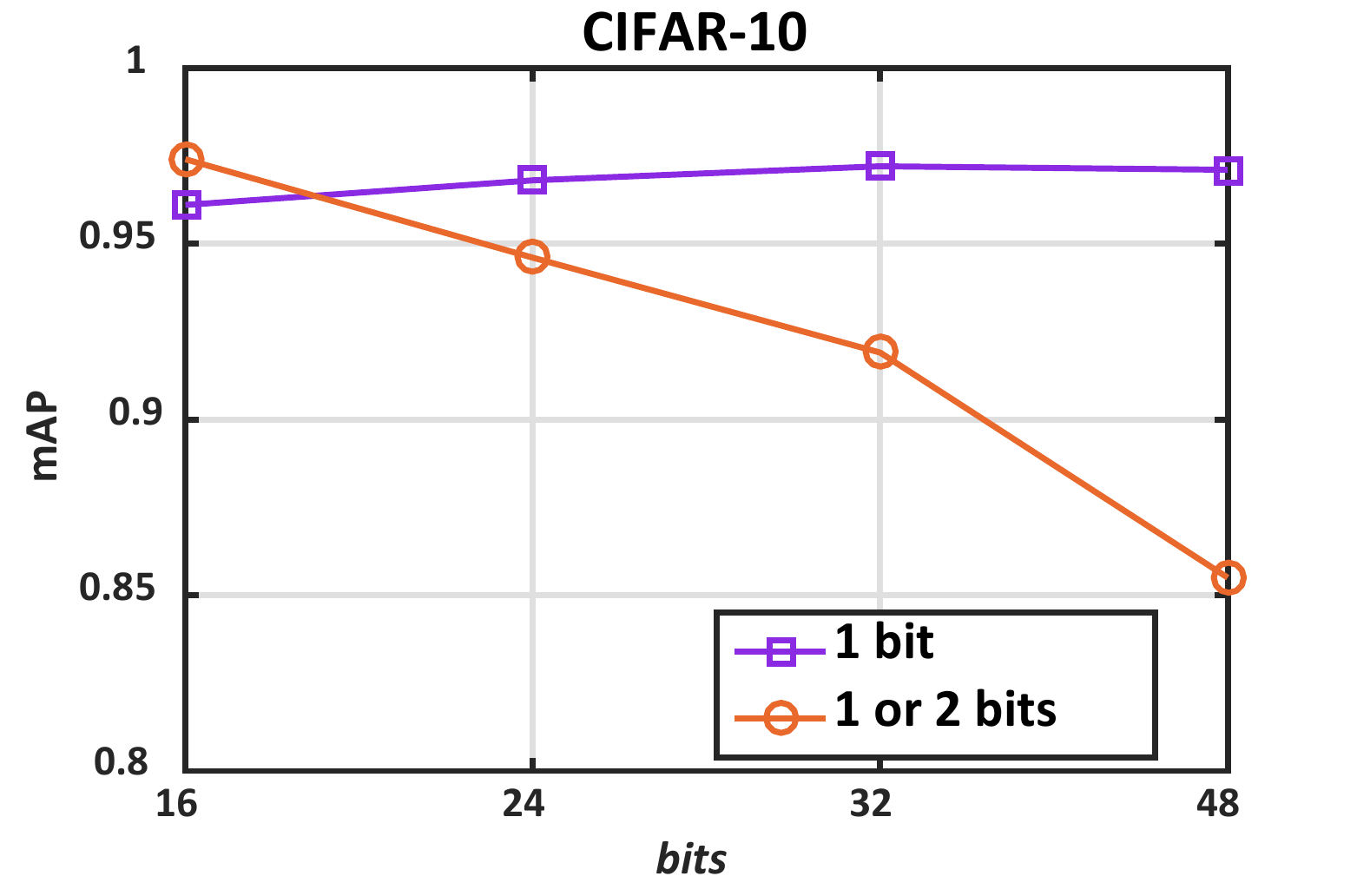}}
		\subfigure{\includegraphics[width=4cm]{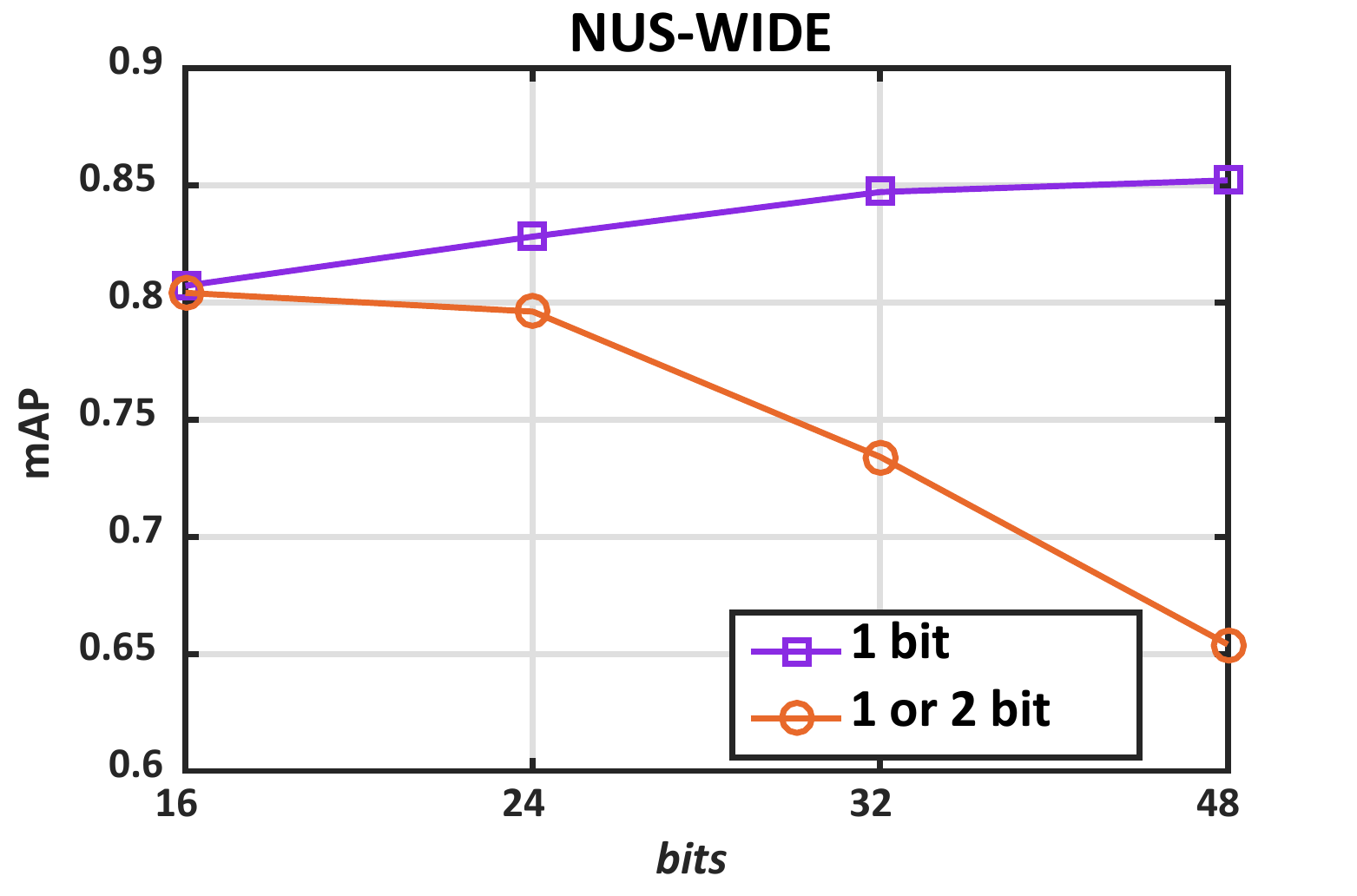}} 
		%\vspace{-2mm}
		\caption{mAP for different action sets.}
		\label{fig_actionset}
		%\vspace{-2mm}
	\end{figure}

	\textbf{Action Set Design.} Our design of action set in Section~\ref{Subsection_Action} is quite straightforward, and here we investigate a more sophisticated design. Instead of allowing only $1$ bit flip, we allow the agent to convert either $1$ bit or $2$ bits by one action. Thus, the size of action set becomes $1 + \mathrm{C}_{b}^{1} + \mathrm{C}_{b}^{2}$, which could be very large for moderate size $b$. As shown in Figure~\ref{fig_actionset}, the performance of $1$-bit action set increases with respect to the number of bits. For $2$-bit action set, its performance is close to that of $1$-bit action set with small $b$, but significantly inferior for relatively large $b$. This is an intuitive phenomenon, because large size of action space makes the true Q-function hard to approximate \cite{sutton1998reinforcement}, thus is usually not favored in reinforcement learning. For example, the possible actions for visual navigation \cite{Zhu_2017_ICCV} are only $4$, which indicates four directions to go. 
	
	Several recent works \cite{Ren_2017_CVPR} have shown that curriculum learning \cite{bengio2009curriculum} could possibly handle the large action space. However, this is beyond the focus of this work, and we leave these questions for future work, \textit{e.g.}, what is the optimal action space, and how to handle large action space in DRLH.
	
	%\subsubsection{Parameter Sensitivity Analysis}
	
	\textbf{Parameter Sensitivity Analysis.} There are two key hyperparameters in our method, $\eta$ in Equation~\ref{Equation_reward_trigger} and the maximum search step $\mathrm{M}$ in the testing phase. In this section, we analyze their sensitivity to validate the robustness of the proposed DRLH.
	
	We use $\mathrm{R_{BCH}}$ to denote the radius of our hyperball based on BCH codes introduced in Section~\ref{Subsection_Reward_Function}. In Table~\ref{sensitive_1}, we show the impact for various $\eta$. As in Equation~\ref{Equation_reward_trigger}, $\eta$ is a hyperparameter that triggers the termination of search, $0 \leq \eta \leq \mathrm{R_{BCH}}$. $\eta = 0$ means that the final reward will not be granted to the agent unless it transforms the binary embedding to the exact location of its class; while $\eta = \mathrm{R_{BCH}}$ means that we just require the agent to put the binary code into the ball of its class. As shown in Table~\ref{sensitive_1} left, the best performance is achieved usually when $\eta$ is around $\frac{1}{2} \mathrm{R_{BCH}}$. As $\eta$ goes down from $\frac{1}{2} \mathrm{R_{BCH}}$ to $0$ or goes up from $\frac{1}{2} \mathrm{R_{BCH}}$ to $\mathrm{R_{BCH}}$, the performance drops monotonically.

	\begin{table}[!t]
		\caption{Evaluation on the impact of hyperparameter $\eta$.}
		\vspace{2mm}
		\centering
		\begin{adjustbox}{width=1.0\columnwidth}
			\begin{tabular}{cc|cccccccc}
				\toprule
				\multicolumn{2}{c|}{$\eta$}           & \multicolumn{1}{c}{0} & \multicolumn{1}{c}{1} & \multicolumn{1}{c}{2} & \multicolumn{1}{c}{3} & \multicolumn{1}{c}{4} & \multicolumn{1}{c}{5} & \multicolumn{1}{c}{6} & \multicolumn{1}{c}{7} \\ \midrule
				\multirow{2}{*}{CIFAR-10} & 16 bits & 0.917                  & 0.945                  & \textbf{0.949}         & 0.932                  & -&-&-&-     \\
				& 32 bits & 0.867                  & 0.903                  & 0.926                  & 0.934                  & 0.947                  & \textbf{0.952}         & 0.948                  & 0.912                  \\ \midrule
				\multirow{2}{*}{NUS-WIDE} & 16 bits & 0.754                  & \textbf{0.807}         & 0.785                  & 0.757&-&-&-&- \\
				& 32 bits & 0.725                  & 0.776                  & 0.815                  & \textbf{0.827}         & 0.784                  & 0.730                  & 0.704                  & 0.648                  \\ \bottomrule
			\end{tabular}
		\end{adjustbox}
		\label{sensitive_1}
	\end{table}
	
	\begin{table}[!t]
		\caption{Evaluation on the impact of hyperparameter $\mathrm{M}$.}
		\vspace{2mm}
		\centering
		\begin{adjustbox}{width=1.0\columnwidth}
			\begin{tabular}{cc|cccccccc}
				\toprule
				\multicolumn{2}{c|}{$\mathrm{M}$}             & \multicolumn{1}{c}{1} & \multicolumn{1}{c}{2} & \multicolumn{1}{c}{4} & \multicolumn{1}{c}{8} & \multicolumn{1}{c}{16} & \multicolumn{1}{c}{32} & \multicolumn{1}{c}{64} & \multicolumn{1}{c}{128} \\ \midrule
				\multirow{2}{*}{CIFAR-10} & 16 bits & 0.614                  & 0.701                  & 0.825                  & 0.922                  & 0.930                   & \textbf{0.949}          & \textbf{0.949}          & \textbf{0.949}           \\
				& 32 bits & 0.440                  & 0.623                  & 0.775                  & 0.897                  & 0.925                   & \textbf{0.948}          & \textbf{0.948}          & \textbf{0.948}           \\ \midrule
				\multirow{2}{*}{NUS-WIDE} & 16 bits & 0.457                  & 0.599                  & 0.678                  & 0.753                  & \textbf{0.807}          & \textbf{0.807}          & \textbf{0.807}          & \textbf{0.807}           \\
				& 32 bits & 0.430                  & 0.527                  & 0.655                  & 0.692                  & 0.743                   & \textbf{0.847} & \textbf{0.847}          & \textbf{0.847}           \\ \bottomrule
			\end{tabular}
		\end{adjustbox}
		
		\label{sensitive_2}
	\end{table}
	
	\begin{figure}[t]
		\begin{center}
			\subfigure[HashNet.]{\includegraphics[width=4cm]{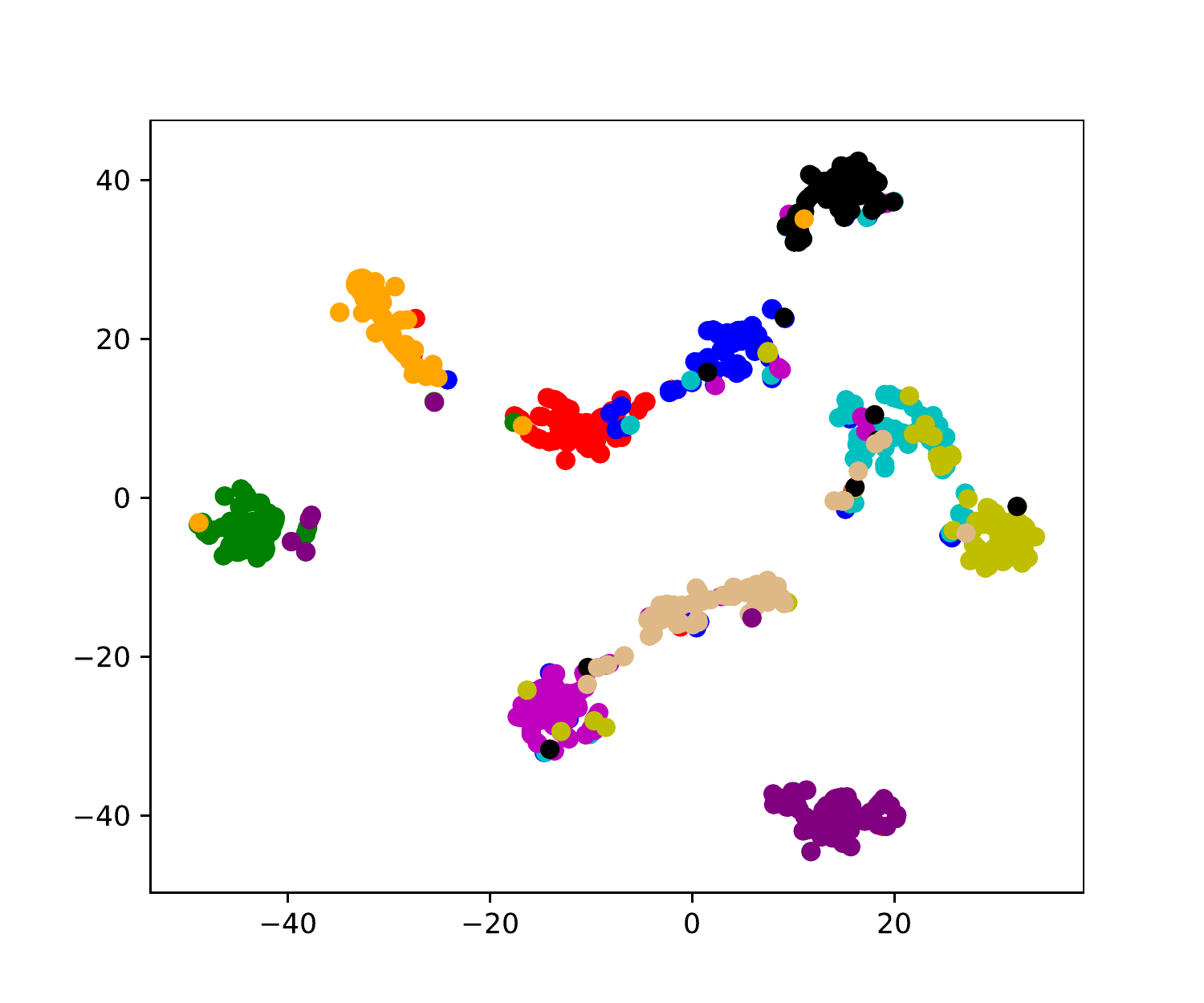}}
			\subfigure[DRLH.]{\includegraphics[width=4cm]{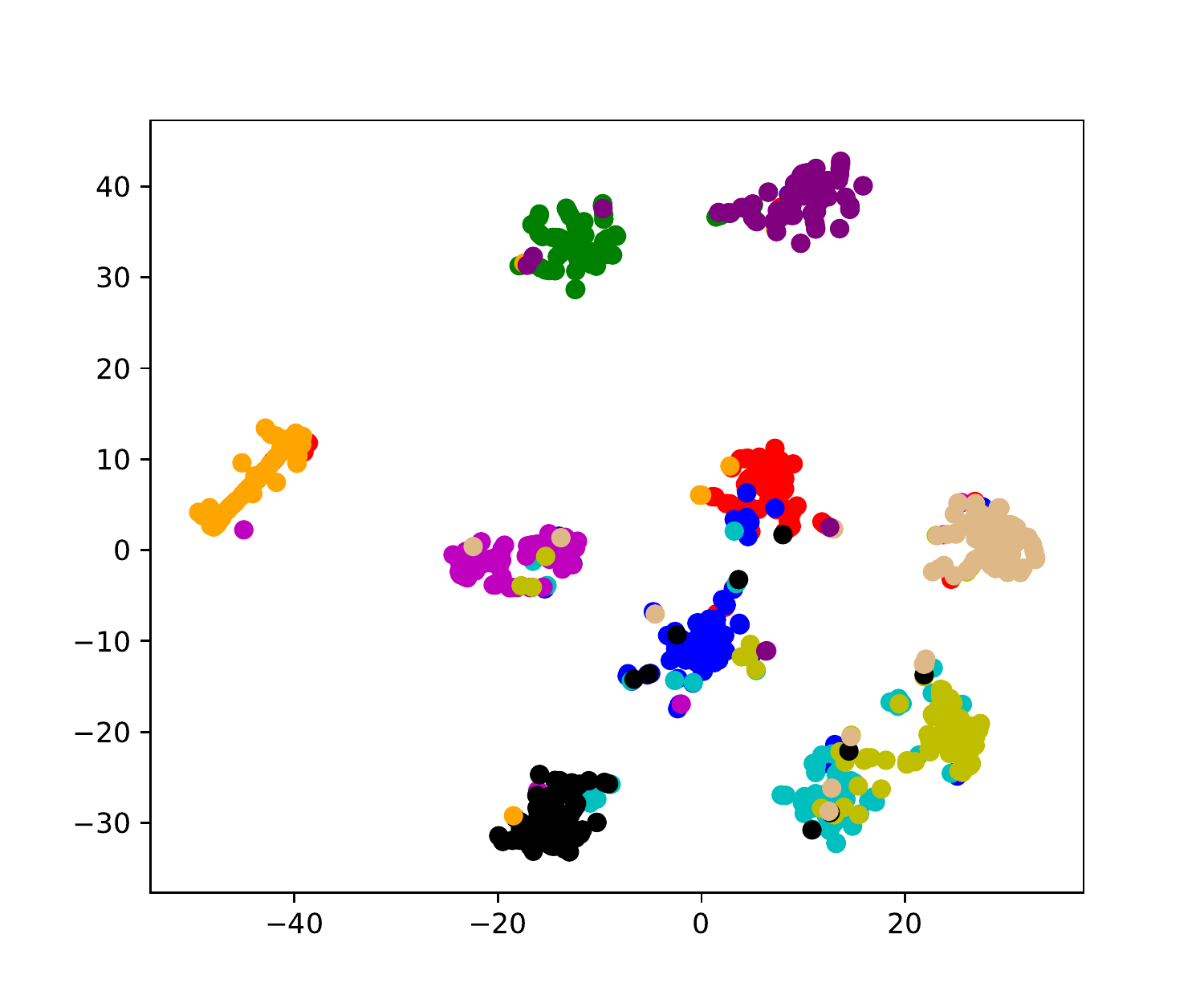}}
		\end{center}
		%\vspace{-2mm}
		\caption{\textbf{Visualization of hash codes using t-SNE.} Our DRLH produces more discriminative embeddings that are intra-class compact and inter-class separable. The two classes relatively close in bottom right are automobile and truck.}
	%	\vspace{-4mm}
		\label{figure_tSNE}
	\end{figure}
	
	In Table~\ref{sensitive_2}, we show the impact of different $\mathrm{M}$ on our model. The performance monotonically increases with respect to $\mathrm{M}$ until the threshold is reached. After $\mathrm{M}$ exceeds the threshold, the agent could either trigger the terminator before the $\mathrm{M}$-th action, or get stuck in a local region. Therefore, a $\mathrm{M}$ larger than the threshold does not lead to better accuracy. Since our model takes around $b$ steps to terminate, we set $\mathrm{M} = b$ in the following experiments.
	
	Also, we visualize the hash codes generated by the HashNet~\cite{cao2017hashnet} and DRLH methods on the CIFAR-10 dataset using t-SNE~\cite{maaten2008visualizing} scheme in Figure~\ref{figure_tSNE}.
	
	\begin{table*}[!t]
		\caption{The mAP on CIFAR-10 dataset and NUS-WIDE dataset using traditional features. The mAP for NUS-WIDE dataset is calculated based on the top 50,00 returned neighbors. We use Italic font to distinguish the unsupervised hashing methods.}
		\vspace{1mm}
		\centering
		\begin{adjustbox}{width=1.8\columnwidth}
			\begin{tabular}{c|cccc|c|cccc}
				\toprule
				\multirow{2}{*}{Method} & \multicolumn{4}{c|}{CIFAR-10}         & \multirow{2}{*}{Method} & \multicolumn{4}{c}{NUS-WIDE}         \\ %\cline{2-5} \cline{7-10} 
				& 12 bits & 24 bits & 32 bits & 48 bits &                         & 12 bits & 24 bits & 32 bits & 48 bits \\ \midrule
				\textit{SH} \cite{weiss2009spectral}                      & 0.127   & 0.128   & 0.126   & 0.129   & \textit{SH} \cite{weiss2009spectral}                      & 0.454   & 0.406   & 0.405   & 0.400   \\ 
				\textit{ITQ} \cite{gong2011iterative}                     & 0.162   & 0.169   & 0.172   & 0.175   & \textit{ITQ} \cite{gong2011iterative}                     & 0.452   & 0.468   & 0.472   & 0.477   \\
				LFH \cite{zhang2014supervised}                     & 0.176   & 0.231   & 0.211   & 0.253   & LFH \cite{zhang2014supervised}                     & 0.571   & 0.568   & 0.568   & 0.585   \\
				KSH \cite{liu2012supervised}                     & 0.303   & 0.337   & 0.346   & 0.356   & KSH \cite{liu2012supervised}                     & 0.556   & 0.572   & 0.581   & 0.588   \\
				SDH \cite{shen2015supervised}                     & 0.285   & 0.329   & 0.341   & 0.356   & SDH \cite{shen2015supervised}                     & 0.568   & 0.600   & 0.608   & 0.637   \\
				FastH \cite{lin2014fast}                   & 0.305   & 0.349   & 0.369   & 0.384   & FastH \cite{lin2014fast}                   & 0.621   & 0.650   & 0.665   & 0.687   \\ \hline
				DRLH (Ours)                    & \textbf{0.556} & \textbf{0.621} & \textbf{0.655} & \textbf{0.671} & DRLH (Ours)                    & \textbf{0.694} & \textbf{0.748} & \textbf{0.768} & \textbf{0.781} \\ \bottomrule
			\end{tabular}
		\end{adjustbox}
		\label{Table_tratitional_method_traditional_feature}
	\end{table*}
	
	\begin{table*}[!t]
		\caption{The mAP on CIFAR-10 dataset and NUS-WIDE dataset using deep features extracted by CNN-F model \cite{chatfield2014return}. The mAP for NUS-WIDE dataset is calculated based on the top 50,00 returned neighbors. We use Italic font to distinguish the unsupervised hashing methods.}
		\vspace{1mm}
		\centering
		\begin{adjustbox}{width=1.8\columnwidth}
			\begin{tabular}{c|cccc|c|cccc}
				\toprule
				\multirow{2}{*}{Method} & \multicolumn{4}{c|}{CIFAR-10}                                     & \multirow{2}{*}{Method} & \multicolumn{4}{c}{NUS-WIDE}                                     \\ %\cline{2-5} \cline{7-10} 
				& 16 bits        & 24 bits        & 32 bits        & 48 bits        &                         & 16 bits        & 24 bits        & 32 bits        & 48 bits        \\ \midrule
				\textit{SH} \cite{weiss2009spectral}               & 0.183          & 0.164          & 0.161          & 0.161          & \textit{SH} \cite{weiss2009spectral}                & 0.621          & 0.616          & 0.615          & 0.612          \\
				\textit{ITQ} \cite{gong2011iterative}               & 0.237          & 0.246          & 0.255          & 0.261          & \textit{ITQ} \cite{gong2011iterative}               & 0.719          & 0.739          & 0.747          & 0.756          \\
				LFH \cite{zhang2014supervised}              & 0.208          & 0.242          & 0.266          & 0.339          & LFH \cite{zhang2014supervised}              & 0.695          & 0.734          & 0.739          & 0.759          \\
				KSH \cite{liu2012supervised}              & 0.488          & 0.539          & 0.548          & 0.563          & KSH \cite{liu2012supervised}              & 0.768          & 0.786          & 0.790          & 0.799          \\
				SDH \cite{shen2015supervised}              & 0.478          & 0.557          & 0.584          & 0.592          & SDH \cite{shen2015supervised}              & \textbf{0.780}          & 0.804          & 0.815          & 0.824          \\
				FastH \cite{lin2014fast}            & 0.553          & 0.607          & 0.619          & 0.636          & FastH \cite{lin2014fast}            & 0.779          & 0.807          & 0.816          & 0.825          \\
				TALR-AP \cite{He2018CVPR} & 0.732 & 0.789 & 0.80 & 0.826 &TALR-AP \cite{He2018CVPR}& 0.709 & 0.734& 0.745& 0.752 \\
				HBMP \cite{Cakir2018ECCV} & \textbf{0.799} &0.804&0.83&0.831&HBMP \cite{Cakir2018ECCV}&0.757&0.805&0.822&\textbf{0.84} \\
				 \midrule
				DRLH (Ours)                    & 0.761 & \textbf{0.812} & \textbf{0.845} & \textbf{0.868} & DRLH (Ours)                    & \textbf{0.780} & \textbf{0.814} & \textbf{0.826} & 0.833 \\ \bottomrule
			\end{tabular}
		\end{adjustbox}
		\label{Table_traditional_method_with_CNN_feature}
	\end{table*}
	
	\subsection{Comparisons with traditional methods}
	\label{Compare_with_traditional_methods}
	
	We follow the settings in the previous works\cite{lai2015simultaneous,li2016feature} to make fair comparison with traditional methods. In CIFAR-10, we randomly select 100 images per class (1,000 images in total) as the test query set, 500 images per class (5,000 images in total) as the training set. For NUS-WIDE dataset, we randomly sample 100 images per class (2,100 images in total) as the test query set, 500 images per class (10,500 images in total) as the training set. The similar pairs are constructed according to the image labels: two images will be considered similar if they share at least one common semantic label. Otherwise, they will be considered dissimilar. Similar to previous works \cite{li2017deep}, we represent each image in CIFAR-10 by a 512-dimensional GIST descriptor \cite{oliva2001modeling}, and each image in NUS-WIDE by a 1134-dimensional feature vector, which comprises color histogram, color auto-correlogram, edge direction histogram, \textit{etc}.
	
	The mAP results of all methods on CIFAR-10 and NUS-WIDE are listed in Table~\ref{Table_tratitional_method_traditional_feature}. The proposed method substantially outperforms the traditional hashing methods on both datasets. In particular, FastH~\cite{lin2014fast} achieves the best performance among all the other methods except for our DRLH on CIFAR-10 dataset. Compared with FastH~\cite{lin2014fast}, our method can improve the performance by at least $25\%$, \textit{e.g.}, from $30\%$ to $55\%$ with $12$ bits. These results verify the advantages of deep hashing over traditional methods. Also, the proposed DRLH achieves superior performance to the best traditional hashing methods on NUS-WIDE dataset.
	
	\subsection{Comparisons with traditional methods using deep features}
	\label{Comparison_with_traditional_methods_using_deep_learned_features}
	
	In order to have a fair comparison, we also compare with traditional hashing methods using deep learned features extracted by the CNN-F network. The mAP results of different methods are listed in Table~\ref{Table_traditional_method_with_CNN_feature}. As shown in Table~\ref{Table_traditional_method_with_CNN_feature}, most of the traditional hashing methods obtain a better retrieval performance using deep learned features. The average mAP results of FastH and SDH on CIFAR-10 dataset are 0.604 and 0.553, respectively. And the average mAP result of our method on CIFAR-10 dataset is 0.821, which outperforms the traditional hashing methods with deep learned features.

	\begin{table*} %[ht]
		\caption{The mAP of different deep hashing methods on CIFAR-10 dataset and NUS-WIDE dataset. The mAP for NUS-WIDE dataset is calculated based on the top 50,000 returned neighbors.}
		\vspace{1mm}
		\centering
		\begin{adjustbox}{width=1.8\columnwidth}
			\begin{tabular}{c|cccc|c|cccc}
				\toprule
				\multirow{2}{*}{Method} & \multicolumn{4}{c|}{CIFAR-10}                                     & \multirow{2}{*}{Method} & \multicolumn{4}{c}{NUS-WIDE}                                     \\ %\cline{2-5} \cline{7-10} 
				& 16 bits        & 24 bits        & 32 bits        & 48 bits        &                         & 16 bits        & 24 bits        & 32 bits        & 48 bits        \\ \midrule
				\textit{GraphBit}~\cite{Duan_2018_CVPR}                    & 0.322 & 0.367 & 0.399 & -   & \textit{GraphBit}~\cite{Duan_2018_CVPR} & -&-&-&-  \\
				DSRH \cite{zhao2015deep}                    & 0.608          & 0.611          & 0.617          & 0.618          & DSRH \cite{zhao2015deep}                   & 0.609          & 0.618          & 0.621          & 0.631          \\
				DQN \cite{cao2016deep}                   & 0.554          & 0.558          & 0.564          & 0.580          & DQN \cite{cao2016deep}                   & 0.768          & 0.776          & 0.783          & 0.792          \\
				NINH \cite{lai2015simultaneous}                    & 0.608          & 0.611          & 0.617          & 0.618          & NINH \cite{lai2015simultaneous}                   & 0.609          & 0.618          & 0.621          & 0.631          \\
				DPSH \cite{li2016feature}                   & 0.763          & 0.781          & 0.795          & 0.807          & DPSH \cite{li2016feature}                  & 0.715          & 0.722          & 0.736          & 0.741          \\
				SSDH \cite{yang2017supervised}                    & 0.897          & 0.898          & 0.899          & 0.900          & SSDH* \cite{yang2017supervised}                   & 0.725          & 0.733          & 0.746          & 0.752          \\
				PGDH~\cite{Yuan_2018_ECCV}                    & 0.736   & 0.741    & 0.747    & 0.762    & PGDH~\cite{Yuan_2018_ECCV}                   & 0.761    & 0.780    & 0.786    & 0.792   \\
				DRLIH~\cite{zhang2018deep}                    & -   & 0.843   & 0.855   & 0.853   & DRLIH~\cite{zhang2018deep}                      & -   & -  &  -  & -   \\
				VDSH \cite{zhang2016efficient}                    & 0.871          & -            & 0.880          & -            & VDSH \cite{zhang2016efficient}                    & -&-&-&-                                          \\
				DTSH \cite{wang2016deep}                    & 0.915          & 0.923          & 0.925          & 0.926          & DTSH \cite{wang2016deep}                   & 0.756          & 0.776          & 0.785          & 0.799          \\
				DSDH~\cite{li2017deep}                    & 0.935   & 0.940   & 0.939   & 0.939   & DSDH~\cite{li2017deep}                    & \textbf{0.815}   & 0.814   & 0.820   & 0.821   \\
				SDSH~\cite{Pidhorskyi1_2018_ACCV}                    & 0.938 & 0.939 & 0.939 & 0.934   & SDSH~\cite{Pidhorskyi1_2018_ACCV}                    & -  & -  & -   & -   \\
				HashNet~\cite{cao2017hashnet}                    & 0.933 & 0.935 & 0.941 & 0.942   & HashNet~\cite{cao2017hashnet} & 0.662 & 0.689 & 0.711 & 0.716  \\
				GreedyHash~\cite{NIPS2018_7360}                    & 0.942 & 0.943 & 0.943 & 0.944   & GreedyHash~\cite{NIPS2018_7360} & -&-&-&-  \\ 
				%TALR-AP \cite{He2018CVPR} & 0.939 & 0.941 & 0.943 & 0.945 %&TALR-AP \cite{He2018CVPR} &0.795&\textbf{0.835}&\textbf{0.848}&\textbf{0.862}\\
				HBMP \cite{Cakir2018ECCV} & 0.942 & 0.944 & 0.945 & 0.945&HBMP \cite{Cakir2018ECCV}& 0.804&0.829&0.841&0.855\\
				DAGH \cite{Chen2019iccv} & 0.934&0.933&0.934&0.932&DAGH \cite{Chen2019iccv}&0.76&0.789&0.793&0.802\\
				\midrule
				Ours                    & \textbf{0.949} & \textbf{0.948} & \textbf{0.952} & \textbf{0.955} & Ours & 0.807 & 0.828 & 0.837& 0.842\\ \bottomrule
			\end{tabular}
		\end{adjustbox}
		\label{Table_deep_hashing}
		%\vspace{-2mm}
	\end{table*}

On NUS-WIDE dataset, our method also achieves better performance compared with traditional hashing methods, as shown in the right part of Table~\ref{Table_traditional_method_with_CNN_feature}. Different from CIFAR-10 dataset, there exits more categories in NUS-WIDE, and each image contains multiple labels. Meanwhile, we only use 500 images per class for training, which may not be enough for DRLH to learn to explore the binary code space. Therefore, our result is only slightly better than SDH~\cite{shen2015supervised}, which is the best among all the compared methods. In Section~\ref{Compare_with_deep_hashing}, we will show that with more training images per class for the NUS-WIDE dataset, our method can achieve far better performances.
	
\subsection{Comparisons with deep hashing}
\label{Compare_with_deep_hashing}
	
	Deep hashing methods usually require many training images to learn the hash function. In this section, we compare with other deep hashing methods under a different experimental setting with more training images. For CIFAR-10, 1,000 images per class are selected as the test query set, the remaining 50,000 images are used as the training set. In NUS-WIDE, 100 images per class are randomly sampled as the test query images, while the remaining 193,734 images are used as the training set. Here we adopt AlexNet to make fair comparisons with prior arts \cite{cao2017hashnet,yang2017supervised}, but more powerful networks like ResNet~\cite{he2016deep} should lead to even better mAP.
	
	Table~\ref{Table_deep_hashing} lists mAP results for different methods under the second experimental setting. As shown in Table~\ref{Table_deep_hashing}, with more training images, our method performs better than that in Section~\ref{Comparison_with_traditional_methods_using_deep_learned_features}. For CIFAR-10 dataset, DSDH and GreedyHash have a significant advantage over other deep hashing methods with the average mAP as 0.938 and 0.943, respectively, while the average mAP result of our method is 0.951, higher than that of DTSH and DSDH.
	
	%\subsubsection{}
	
\begin{figure}[!t]
		\centering
		\subfigure[Total training time.]{\includegraphics[width=4cm]{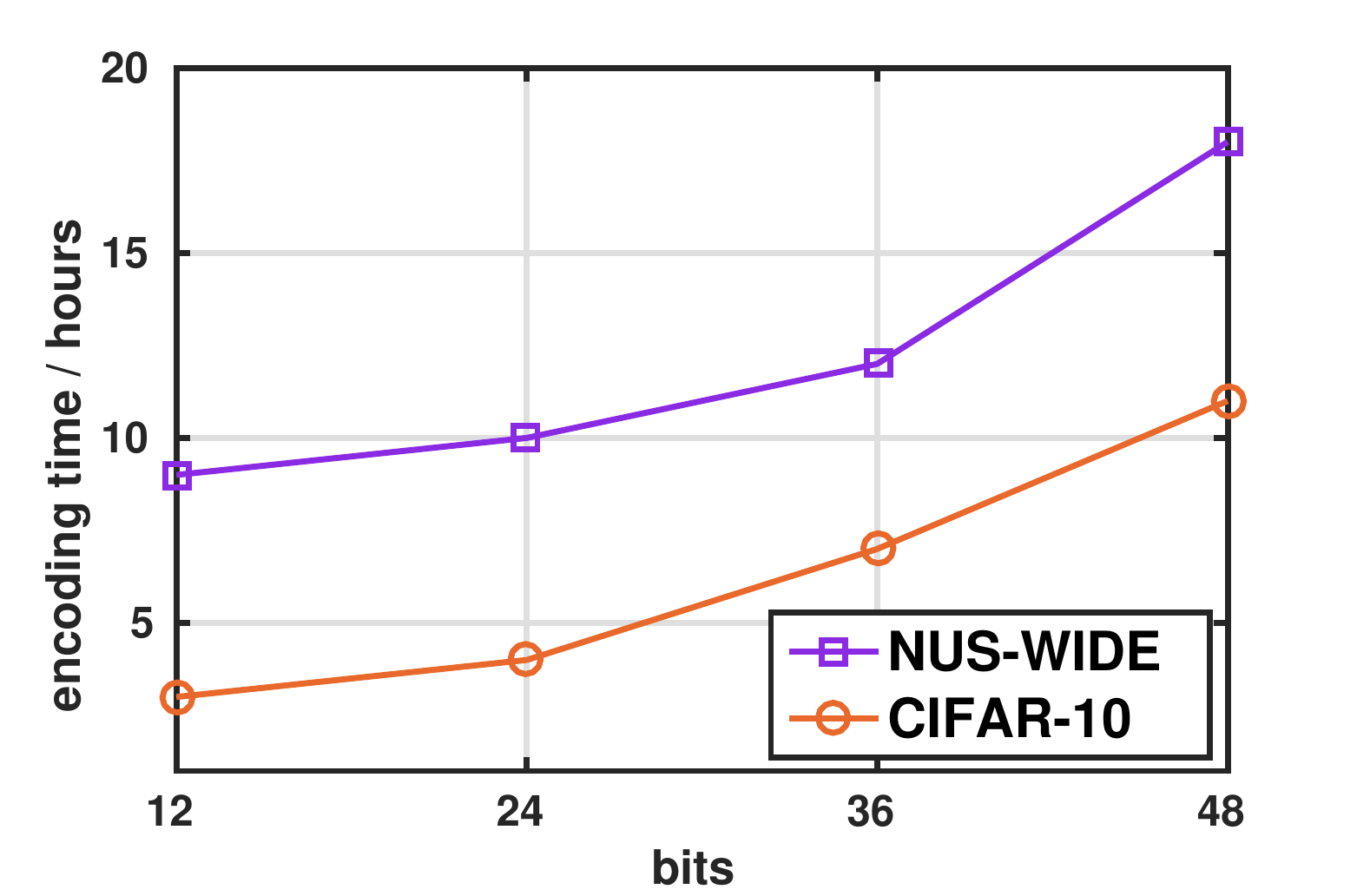}}
		\subfigure[Average inference time.]{\includegraphics[width=4cm]{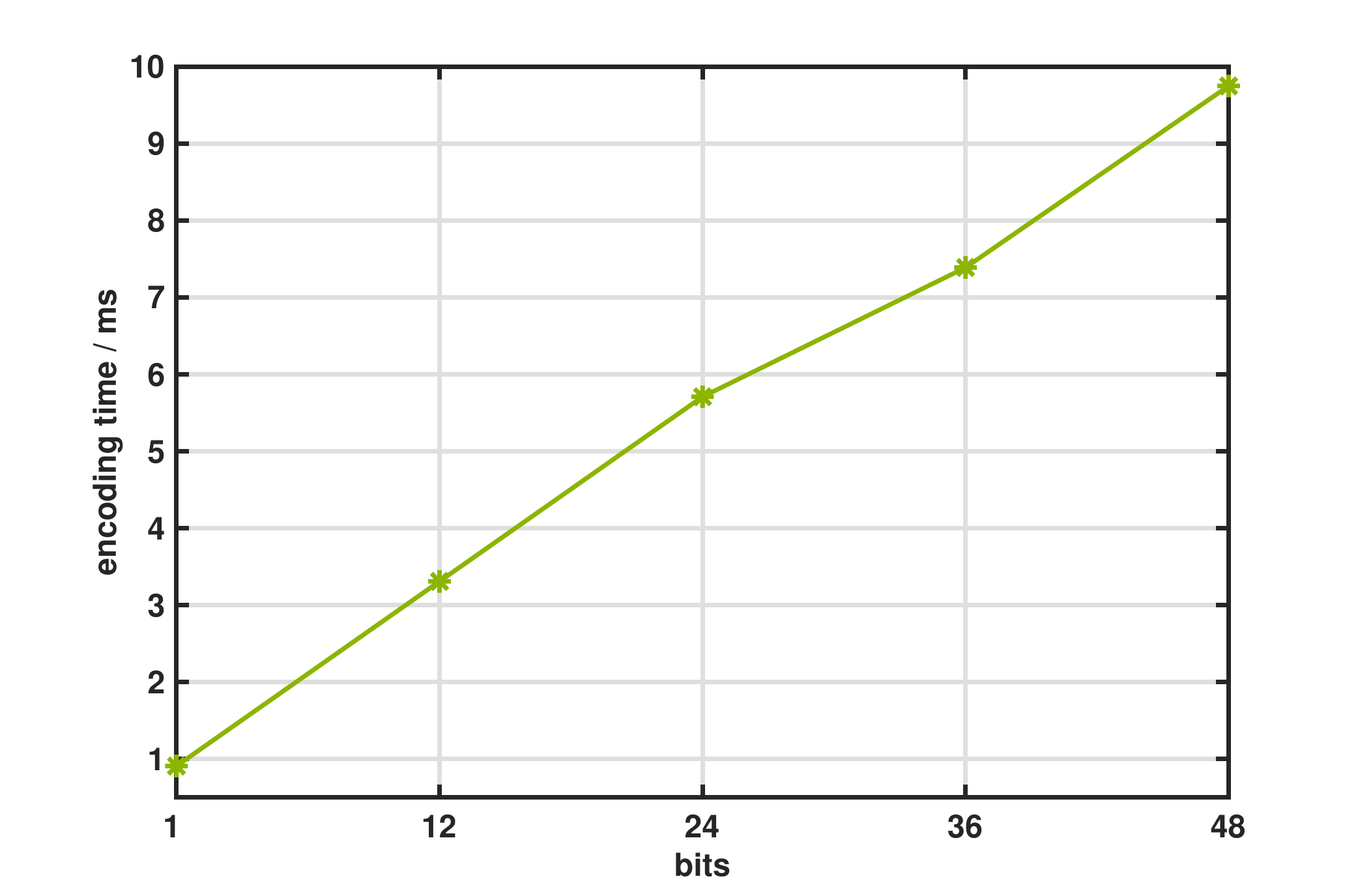}}
		%\vspace{-5mm}
		\caption{Train/Test time of our proposed DRLH.}
		\label{figure_training_time}
	\end{figure}

	\textbf{Timing.} DRLH is fast to train. In case of 48 bits using 8 GPUs, DRLH takes 17 hours to train on CIFAR-10, and 28 hours on NUS-WIDE. The training time for other code length is shown in Figure~\ref{figure_training_time}. During testing, the computation of DRLH consists of two parts: the feature extraction by CNN, and the decision making by Q-network. Compared with existing deep hashing works like \cite{yang2017supervised,cao2017hashnet}, the primary overhead of DRLH is the multiple running of the Q-network for sequential decision making, which is mainly conditioned on the code length. Thus, we measure the inference time \textit{w.r.t} the code length and present the results in Figure~\ref{figure_training_time}. As expected, the inference time increases linearly to the number of bits. Also, the time cost for changing 1 bit can be roughly regarded as the encoding time for most existing deep hashing work like \cite{yang2017supervised,cao2017hashnet}, which usually append a binary encoding layers to the CNN feature extractor. Note that in practical image retrieval application, the inference time is usually not the bottleneck, compared with the linear scan over the entire database. Therefore, we consider our inference time acceptable. We note that the memory cost and computational overhead of the fc layers in our Q-network could be partly eliminated by Fastfood transform \cite{yang2015deep} or TT-format \cite{novikov2015tensorizing} when necessary.
	
	\section{Conclusion}
	
	In this work, we present a novel decision-making approach for deep supervised hashing, which achieves state-of-the-art performance on two benchmark datasets. Different from previous hashing methods, we consider the binary embedding problem as travelling in the binary code space, and utilize a deep Q-network to figure out a path in the binary space to reach the best binary embedding. To learn our Q-network, we design policy gradient approach with BCH codes based rewards. We conduct detailed analysis and extensive ablation studies on our approach to understand its merits and properties. Our future works include investigating the inter-class variances and taking the correlations between labels into consideration to design better reward scheme.

%\clearpage

\bibliographystyle{ieee_fullname}
\bibliography{DRLHashing}
\end{document}